\documentclass{article}
\usepackage{spconf,amsmath,graphicx}

\usepackage{graphicx}
\usepackage{amsmath}
\usepackage{amssymb}
\usepackage{booktabs}
\usepackage{multirow}
\usepackage{enumitem}
\usepackage{subfloat}
\usepackage{subcaption}
\usepackage{bm}

\usepackage{amsmath,amsfonts}
\usepackage{algorithmic}
\usepackage{array}
\usepackage{textcomp}
\usepackage{stfloats}
\usepackage{verbatim}

\usepackage{multirow}
\usepackage{booktabs} 
\usepackage{makecell}
\usepackage{amssymb}
\usepackage{times}
\usepackage{soul}
\usepackage{url}
\usepackage[hidelinks]{hyperref}
\usepackage[utf8]{inputenc}
\usepackage{float}
\usepackage{amsmath}
\usepackage{multirow}
\usepackage{amsthm}
\usepackage{balance}
\usepackage{booktabs}
\usepackage{algorithm}
\usepackage{algorithmic}
\usepackage{pbox}

\usepackage{amsmath}
\usepackage{amssymb}
\usepackage{booktabs}
\usepackage{caption}
\usepackage{tikz}

\title{AxWin Transformer: A Context-Aware Vision Transformer Backbone with Axial Windows}
\name{Fangjian Lin$^{1,2}$, Yizhe Ma$^{1}$, Sitong Wu$^{3}$, Long Yu$^{1}$, ShengWei Tian$^{1\dag}$\thanks{$\dag$ Corresponding author.}}
\address{School of Software, Xinjiang University, Urumqi, China$^1$\\ Shanghai AI Laboratory, Shanghai China$^2$\\ University of Chinese Academy of Sciences$^3$}
\begin{document}

\maketitle

\begin{abstract}
Recently Transformer has shown good performance in several vision tasks due to its powerful modeling capabilities. To reduce the quadratic complexity caused by the attention, some outstanding work restricts attention to local regions or extends axial interactions. However, these methos often lack the interaction of local and global information, balancing coarse and fine-grained information. To address this problem, we propose AxWin Attention, which models context information in both local windows and axial views. Based on the AxWin Attention, we develop a context-aware vision transformer backbone, named AxWin Transformer, which outperforming the state-of-the-art methods in both classification and downstream segmentation and detection tasks.
\end{abstract}


\section{Introduction}

Recently, Transformer has shown remarkable potential in computer vision. Since Dosovitskiy \MakeLowercase{\textit{et al.}} \cite{ViT} proposed Vision Transformer (ViT), the design of the Attention module has become one of the main research hotspots. Several works \cite{PVT, msg, focaltrans, segformer, pit, Swin, CSWin} have achieved high accuracy on classification tasks, but the performance on downstream tasks, especially dense prediction tasks, has not been the same. For dense prediction tasks that include complex scene changes, there is required to have two properties for the backbone: 1. long-range global modeling capability. 2. excellent local information extraction capability. The former not only models rich context information but also obtains higher shape bias (i.e., Some work \cite{RepLKNet, StructToken} has demonstrated that shape bias is critical for downstream tasks). the latter enables the model to focus on the key regions of the feature map. But, balancing these two aspects is a very challenging task.

\begin{figure}[tp]
\centering
\includegraphics[width=1\linewidth]{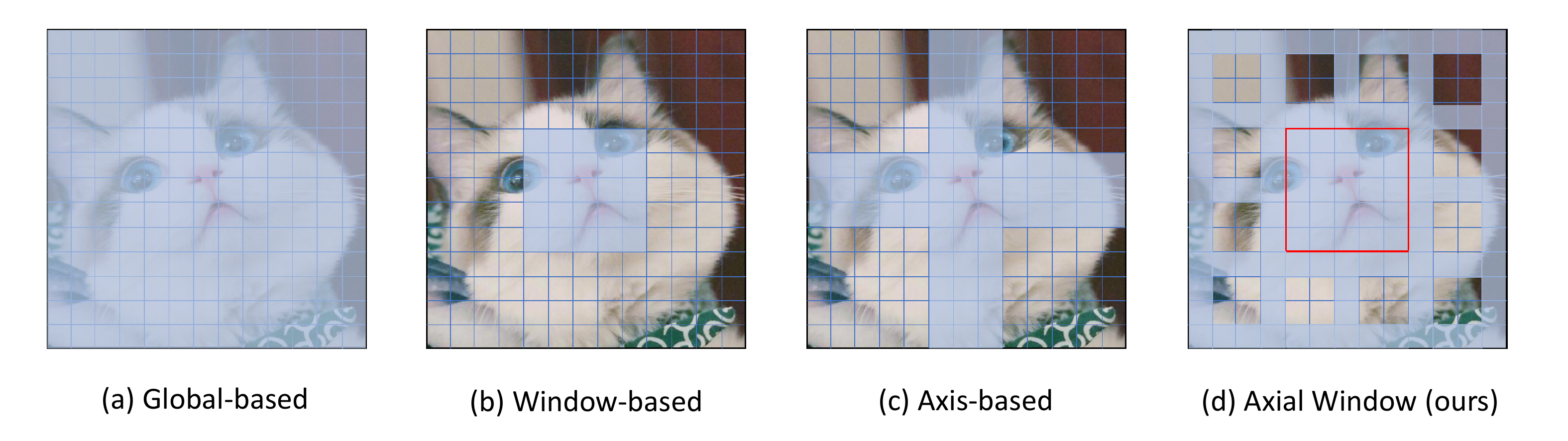} 
\caption{
Comparison with different self-attention mechanisms in Transformer backbones.
The blue area indicates performing attention operation.
(a) is the standard global self-attention.
(b) is the window-based self-attention. It restricts the computation of attention to the inside of each window.
(c) is the Axis-based self-attention. It expands the receptive fields by alternating rows and columns.
(d) Ours, it expands the receptive fields by alternating single rows and columns, and adds windows to focus on local features.
}
\label{diff} 
\end{figure}

For the global modeling capability of the attention, a classical representation is Vision Transformer (ViT) \cite{ViT}. As shown in Figure~\ref{diff} (a), it can model every pixel in an image. However, for downstream tasks where images have high resolution, the quadratic complexity of global self-attention is unbearable on the one hand, and it lacks local modeling capability on the other. One way to solve the quadratic complexity of global self-attention is to divide the global image into multiple local regions, as shown in Figure~\ref{diff} (b), Swin Transformer \cite{Swin} reduces the computational complexity to a tolerable level and enables the model to focus its attention on the regions inside each window. Although its shift transform can expand the receptive fields, this operation is not sufficient for global dependencies. Axial-based methods are a more friendly choice for downstream tasks, where axial attention can obtain higher shape bias and greatly reduce computational complexity. As shown in Figure~\ref{diff} (c), such as CSWin Transformer \cite{CSWin} or Pale Transformer \cite{Pale}, global context interactions are constructed by alternating rows and columns. However, as the image resolution increases in the downstream task, the blank portion of the row and column alternation increases, which may lack some critical information in the image. In short, axial-based methods lack the local modeling capability to capture effective context information.

\begin{figure*}[tp]
\centering
\includegraphics[width=0.7\linewidth]{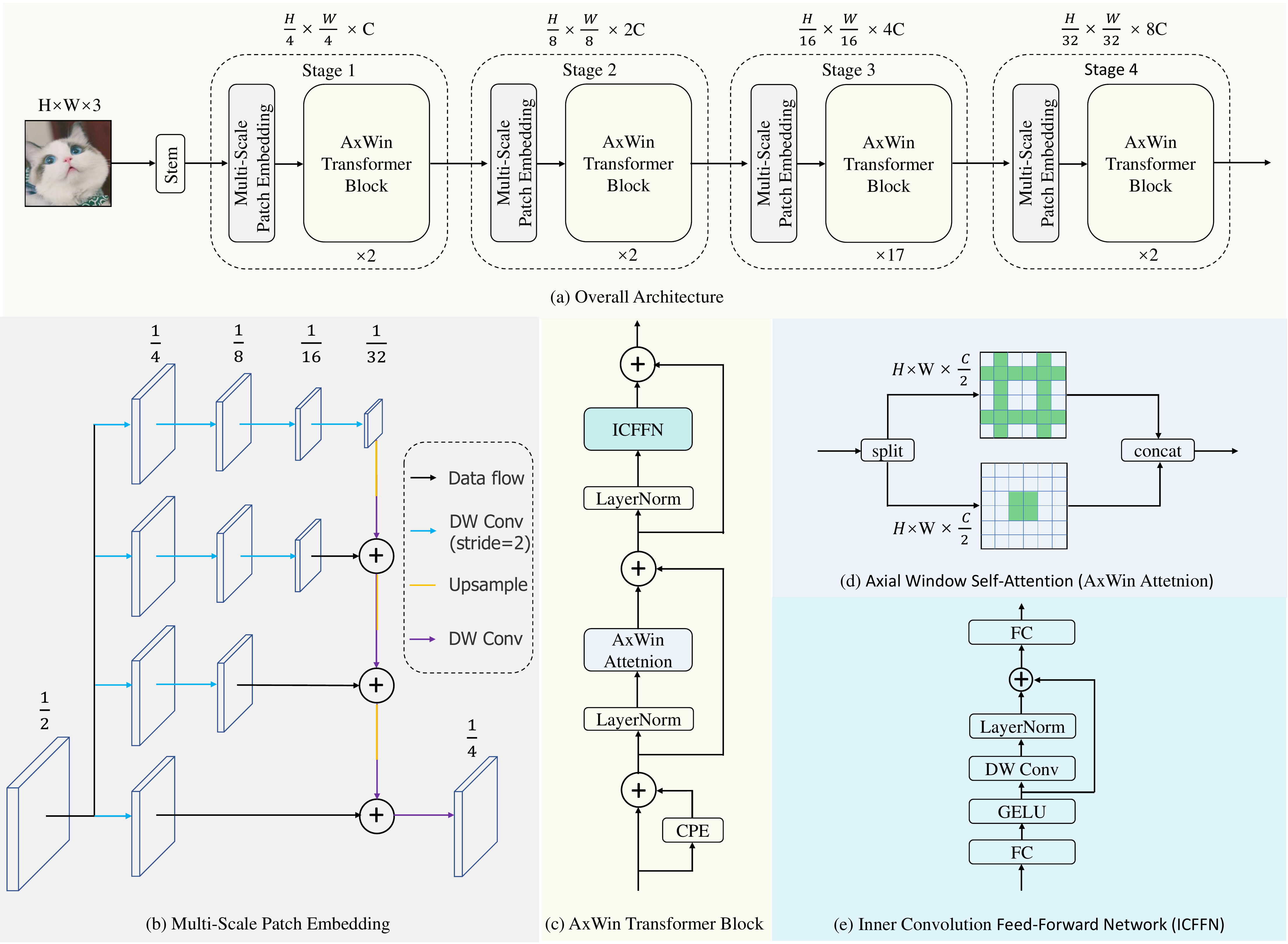} 
\caption{
(a) The overall architecture of our AxWin Transformer.
(b) Illustration of Multi-Scale Patch Embedding downsampling strategy.
(c) The composition of each block.
(d) The specific implementation of AxWin Attention.
(e) Structure of Inner Convolution Feed-Forward Network.
}
\label{framework} 
\end{figure*}
In this work, we propose an axial window of self-attention to solve the above problem, window region for focusing on local features and axial region for modeling global features to obtain higher shape bias and capture richer context dependencies. In addition, we devise a Multi-Scale Patch Embedding (MSPE) downsampling strategy to enrich the context information of high-resolution images. Finally, we slightly improve the classical MLP architecture with a built-in depth-wise convolution operation to enhance the local bias. The architecture of the whole network and the details of each module are shown in Figure \ref{framework}.


\section{related work}
Recent Vision Transformer backbones focus on two main aspects: (1) Enhanced local modeling capabilities. (2) Efficient global attention implementation.

\noindent\textbf{Windows-Based Attention}.
The classical ViT architecture uses global attention and lacks local inductive bias. Swin Transformer\cite{Swin} enhances the ability to extract local information and greatly reduces the computation of self-attention by confining the attention within the window. Consequently, T2T-ViT\cite{T2TVIT}, Shuffle Transformer\cite{shuffle} facilitates the development of Window-based attention by enhancing local connections across windows.

\noindent\textbf{Efficent Global Attention}.
CSWin\cite{CSWin} enhances global context awareness and image shape bias by using axial-based attention to establish global connections across rows and columns. Pale Transformer extends attention to multiple rows and columns, balancing performance and efficiency.

Different from the above two attention mechanisms, our attention module integrates both local and global attention to overcome the problem of insufficient local information extraction and limited global representation.

\section{Method}
In this section, we first show the Multi-Scale Patch Embedding (MSPE) module. Then we describe the efficient implementation of AxWin Attention. Finally the overall architecture of AxWin Transformer and various variant configurations are shown.

\noindent\textbf{Multi-Scale Patch Embedding}.
In order to capture multi-scale context information, we propose the MSPE module.
As shown in Figure 2(a), given an input feature map $X\in\mathcal{R}^{h \times w \times c}$, the output feature map $Y \in\mathcal{R}^{\frac{h}{2} \times \frac{w}{2} \times 2c}$. Corresponding to stages 1-4, the number of branches in MSPE is 4-1.
For branch i, i light-weight 3$\times$3 depth-wise convolutions with stride $=$ 2 are performed, the output feature map $X_i \in\mathcal{R}^{\frac{h}{2^i} \times \frac{w}{2^i} \times 2c}$. For branch i and i+1, A top-down connect operation is used to fuse multi-scale features (i.e., bilinear interpolation to upsample the low-scale feature map), followed by a 3$\times$3 depth-wise convolution with stride $=$ 1 and a 1$\times$1 convolution.

\noindent\textbf{Axial Window Self-Attention}.
In order to capture both fine-grained local features and coarse-grained global information, we propose the Axial Window Self-Attention (AxWin Attention), which computes self-attention within an axial window region. 
As shown in the green shadow of Figure \ref{framework}(d), given an input feature $X\in\mathcal{R}^{h \times w \times c}$, first the fully connected layer is used to perform the mapping of $X$ to generate the Query ($X_q$), Key ($X_k)$) and Value ($X_v$). Then $\{X_q, X_K, X_v\}$ is divided into window group $\{X_{wq}, X_{wk}, X_{wv}\} \in \mathcal{R}^{h \times w \times \frac{c}{2}}$ and axial group $\{X_{aq}, X_{ak}, X_{av}\} \in \mathcal{R}^{h \times w \times \frac{c}{2}}$ according to channel dimension. For the window group, the matrix $\{X_{wq}, X_{wk}, X_{wv}\}$ is split in a non-overlapping manner \cite{Swin}, then perform multi-head self-attention. For the axial group, refer to previous work\cite{Pale}, We rearrange $\{X_{aq}, X_{ak}, X_{av}\}$ into two separate regions of rows $\{X_{aq}^r, X_{ak}^r, X_{av}^r\} \in \mathcal{R}^{s_{ar} \times w \times \frac{c}{4}}$ and columns $\{X_{aq}^c, X_{ak}^c, X_{av}^c\}  \in \mathcal{R}^{h \times s_{ac} \times \frac{c}{4}}$. Here $s_{ar} = s_{ac}$ and indicates how many alternating rows and columns. We perform the connection operation along channel dimension after the division is complete, we get $\{X_{aq}^{rc}, X_{ak}^{rc}, X_{av}^{rc}\}$ to perform multi-head self-attention (MHSA)\cite{ViT}.
\begin{align}
& \tilde{X}_{rc} = MHSA(X_{aq}^{rc}, X_{ak}^{rc}, X_{av}^{rc}),\\
& \tilde{X}_{w} = MHSA(X_{wq}, X_{wk}, X_{wv}), \\
& \hat{X} = Concat(\tilde{X}_{rc}, \tilde{X}_{w}).
\end{align}
\begin{table}[h]
\caption{Detailed configurations of AxWin Transformer variants. $P_i$ means the spatial reduction factor. $C_i$ is the channel dimension. $H_i$, $S_i$ and $R_i$ represent the number of heads, the split-size (i.e., window size and row-column size) for AxWin-Attention, and the expand ratio in ICFFN.}
    \resizebox{1.0\columnwidth}{!}{
        \begin{tabular}{c|c|c|c|c}
            \toprule [1pt]
            \begin{tabular}{c}Stage/Stride\end{tabular}
               & Layer
               & AxWin-T
               & AxWin-S
               & AxWin-B
            \\
            \hline
            Stride=2
               & Stem
               & $P_0=2$, $C_0=32$
               & $P_0=2$, $C_0=48$
               & $P_0=2$, $C_0=56$
            \\
            \hline

            \multirow{4}{*}{ \begin{tabular}{c} Stage 1 \\ Stride=4 \end{tabular}}
               & \begin{tabular}{c}MSPE\end{tabular}
               & $P_1=2$, $C_1=64$
               & $P_1=2$, $C_1=96$
               & $P_1=2$, $C_1=112$
            \\
            \cline{2-5}
               & \begin{tabular}{c}AxWin\\Block\end{tabular}
               & $\begin{bmatrix}\setlength{\arraycolsep}{1pt} \begin{array}{c}
                        H_1$=$2  \\
                        S_1$=$7 \\
                        R_1$=$4
                    \end{array} \end{bmatrix} \times 2$
               & $\begin{bmatrix}\setlength{\arraycolsep}{1pt} \begin{array}{c}
                        H_1$=$2  \\
                        S_1$=$7 \\
                        R_1$=$4
                    \end{array} \end{bmatrix} \times 2$
               & $\begin{bmatrix}\setlength{\arraycolsep}{1pt} \begin{array}{c}
                        H_1$=$4  \\
                        S_1$=$12 \\
                        R_1$=$4
                    \end{array} \end{bmatrix} \times 2$
            \\
            \hline
            \multirow{4}{*}{ \begin{tabular}{c} Stage 2 \\ Stride=8 \end{tabular}}

               & \begin{tabular}{c}MSPE\end{tabular}

               & $P_2=2$, $C_2=128$
               & $P_2=2$, $C_2=192$
               & $P_2=2$, $C_2=224$
            \\
            \cline{2-5}
               & \begin{tabular}{c}AxWin \\Block\end{tabular}
               & $\begin{bmatrix}\setlength{\arraycolsep}{1pt} \begin{array}{c}
                        H_2$=$4  \\
                        S_2$=$7 \\
                        R_2$=$4
                    \end{array} \end{bmatrix} \times 2$
               & $\begin{bmatrix}\setlength{\arraycolsep}{1pt} \begin{array}{c}
                        H_2$=$4  \\
                        S_2$=$7 \\
                        R_2$=$4
                    \end{array} \end{bmatrix} \times 2$
               & $\begin{bmatrix}\setlength{\arraycolsep}{1pt} \begin{array}{c}
                        H_2$=$8  \\
                        S_2$=$12 \\
                        R_2$=$4
                    \end{array} \end{bmatrix} \times 2$
            \\
            \hline
            \multirow{4}{*}{ \begin{tabular}{c} Stage 3 \\ Stride=16 \end{tabular}}

               & \begin{tabular}{c}MSPE\end{tabular}

               & $P_3=2$, $C_3=256$
               & $P_3=2$, $C_3=384$
               & $P_3=2$, $C_3=448$
            \\

            \cline{2-5}
               & \begin{tabular}{c}AxWin \\Block\end{tabular}
               & $\begin{bmatrix}\setlength{\arraycolsep}{1pt} \begin{array}{c}
                        H_3$=$8  \\
                        S_3$=$7 \\
                        R_3$=$4
                    \end{array} \end{bmatrix} \times 17$
               & $\begin{bmatrix}\setlength{\arraycolsep}{1pt} \begin{array}{c}
                        H_3$=$8 \\
                        S_3$=$7 \\
                        R_3$=$4
                    \end{array} \end{bmatrix} \times 17$
               & $\begin{bmatrix}\setlength{\arraycolsep}{1pt} \begin{array}{c}
                        H_3$=$16 \\
                        S_3$=$12 \\
                        R_3$=$4
                    \end{array} \end{bmatrix} \times 17$
            \\

            \hline
            \multirow{4}{*}{ \begin{tabular}{c} Stage 4 \\ Stride=32 \end{tabular}}

               & \begin{tabular}{c}MSPE\end{tabular}

               & $P_4=2$, $C_4=512$
               & $P_4=2$, $C_4=768$
               & $P_4=2$, $C_4=896$
            \\

            \cline{2-5}
               & \begin{tabular}{c}AxWin \\Block\end{tabular}
               & $\begin{bmatrix}\setlength{\arraycolsep}{1pt} \begin{array}{c}
                        H_4$=$16 \\
                        S_4$=$7 \\
                        R_4$=$4
                    \end{array} \end{bmatrix} \times 2$
               & $\begin{bmatrix}\setlength{\arraycolsep}{1pt} \begin{array}{c}
                        H_4$=$16 \\
                        S_4$=$7 \\
                        R_4$=$4
                    \end{array} \end{bmatrix} \times 2$
               & $\begin{bmatrix}\setlength{\arraycolsep}{1pt} \begin{array}{c}
                        H_4$=$32 \\
                        S_4$=$12 \\
                        R_4$=$4
                    \end{array} \end{bmatrix} \times 2$
            \\

            \bottomrule[1pt]
        \end{tabular}
    }
    \label{tab:arch}

\end{table}
\noindent\textbf{AxWin Transformer Block}.
As shown in Figure \ref{framework}(c), there are three main modules in our AxWin Transformer block: the conditional position encoding (CPE), AxWin Attention and Inner Convolution Feed-Forward Network (ICFFN). The CPE\cite{cpe} is used to dynamically generate implicit position embedding, our AxWin Attention is used to capture local and global context information, the proposed ICFFN module is based on the MLP module (i.e., consists of two fully connected layers) with a 3x3 depth-wise convolution to add local information extraction capability for feature projection. The forward process is as follows:
\begin{align}
& \tilde{X}^i = X^{i-1} + \text{CPE}({X^{i-1}})
 \label{eq_CPE},
 \\
& \hat{X}^i = \tilde{X}^i + \text{AxWin Attention}\Big(\text{LN}(\tilde{X}^i)\Big) 
\label{eq:attention},
\\
& X^i = \hat{X}^i + \text{ICFFN \Big(LN}(\hat{X}^i)\Big).
\label{icffn}
\end{align}
We use layer normalization (LN) for feature normalization.

\noindent\textbf{Overall Architecture and Variants}.
Our AxWin Transformer consists of a stem layer, four hierarchical stages, and a classifier head. As shown in Figure \ref{framework} (a), the stem layer\cite{cmt} (i.e., a 3×3 convolution layer with stride = 2 and two 3×3 convolution layers with stride = 1) makes the output features smoother. After the stem, each stage contains a MSPE module and multiple AxWin Transformer blocks. The final classifier head is a linear
layer.
There are three different variants, including AxWin-T (Tiny), AxWin-S (small), and AxWin-B (base), whose detailed configurations are shown in Table \ref{tab:arch}. The above variants differ primarily in the channel dimension and the number of heads.

\begin{table}[h]
\caption{\label{CLS_EXP} Comparisons of different backbones on ImageNet1K validation set. Avg-improve represents the average performance improvement per variant (T, S, B). Flops are calculated with the resolution of 224×224.}
\centering
\resizebox{0.8\columnwidth}{!}{
\begin{tabular}{l|cc|c}
\toprule[1pt]
{Method}         & {Params}    & {FLOPs}   & {Top-1 Acc. (\%)} \\
\midrule[1pt] 
PVT-S \cite{PVT}        & 25M     & 3.8G     & 79.8     \\
Swin-T \cite{Swin}      & 29M     & 4.5G     & 81.3     \\
CSWin-T \cite{CSWin}    & 23M     & 4.3G     & 82.7     \\
Pale-T \cite{Pale}      & 22M     & 4.2G     & 83.4     \\
\textbf{AxWin-T}(ours)  & 22M     & 3.5G     & \textbf{83.9}     \\
\midrule[0.5pt]
PVT-M \cite{PVT}        & 44M     & 6.7G     & 81.2     \\
Swin-S \cite{Swin}      & 50M     & 8.7G     & 83.0     \\
CSWin-S \cite{CSWin}    & 35M     & 6.9G     & 83.6     \\
Pale-S \cite{Pale}      & 48M     & 9.0G     & 84.3     \\
\textbf{AxWin-S}(ours)  & 48M     & 7.6G     & \textbf{84.6}     \\
\midrule[0.5pt]
Swin-B \cite{Swin}      & 88M     & 15.4G    & 83.3     \\
CSWin-B \cite{CSWin}    & 78M     & 15.0G    & 84.2     \\
Pale-B \cite{Pale}      & 85M     & 15.6G    & 84.9     \\
\textbf{AxWin-B}(ours)  & 84M     & 12.7G    & \textbf{85.1}     \\
\bottomrule[1pt]
\end{tabular}}
\end{table}

\section{Experiments}
We first compare our AxWin Transformer with the state-of-the-art methods on ImageNet-1K \cite{imagenet} for image classification. To demonstrate the generalization of our method, we performed experiments on several downstream tasks, including ADE20k \cite{ADE20K} for semantic segmentation, COCO \cite{coco} for object detection, and instance segmentation. Finally, we give the analysis of ablation studies for each module.

\noindent\textbf{Image Classification on ImageNet-1K}.
Table \ref{CLS_EXP} compares the performance of our AxWin Transformer with the state-of-the-art methods on ImageNet-1K validation set. Our method boosts the top-1 accuracy by an average of $1.5\%$ for all variants compared to the most relevant sota methods.


\begin{table}[h]
\small
\caption{\label{EXP_coco} Comparison of different backbones on COCO val2017 using Mask R-CNN framework, and 1x training schedule for object detection and instance segmentation. Flops are calculated with a resolution of 800$\times$1280.}
\addtolength{\tabcolsep}{-4.9pt}
\begin{tabular}{l|cc|cccccc}
\toprule[1pt]
{Backbone}   & {Params}   & {FLOPs}   & AP$^{\text{box}}$    & AP$_{50}^{\text{box}}$     & AP$_{75}^{\text{box}}$    & AP$^{\text{mask}}$    & AP$_{50}^{\text{mask}}$    & AP$_{75}^{\text{mask}}$   \\ 
\midrule[1pt]
PVT-S   \cite{PVT}      & 44M        & 245G    & 40.4  & 62.9  & 43.8  & 37.8   & 60.1  & 40.3 \\
Swin-T  \cite{Swin}     & 48M        & 264G    & 43.7  & 66.6  & 47.6  & 39.8   & 63.3  & 42.7 \\
CSWin-T \cite{CSWin}    & 42M        & 279G    & 46.7  & 68.6  & 51.3  & 42.2   & 65.6  & 45.4 \\
Pale-T  \cite{Pale}     & 41M        & 306G    & 47.4  & 69.2  & 52.3  & 42.7   & 66.3  & 46.2   \\
AxWin-T                 & 41M        & 236G     & \textbf{48.2}  & \textbf{69.7}  & \textbf{52.9}  & \textbf{43.4}  & \textbf{66.9}  & \textbf{46.7}   \\
\midrule[0.5pt]
PVT-M \cite{PVT}        & 64M        & 302G    & 42.0      & 64.4    & 45.6     & 39.0      & 61.6      & 42.1      \\
CSWin-S \cite{CSWin}    & 54M        & 342G    & 47.9      & 70.1    & 52.6     & 43.2      & 67.1      & 46.2      \\ 
Pale-S\cite{Pale}       & 68M        & 432G    & 48.4      & 70.4    & 53.2     & 43.7      & 67.7      & 47.1   \\
AxWin-S                 & 68M        & 318G     & \textbf{49.1}  & \textbf{71.0}  & \textbf{53.9}  & \textbf{44.3}  & \textbf{68.2}  & \textbf{47.5}   \\
\midrule[0.5pt]
PVT-L \cite{PVT}        & 81M      & 364G       & 42.9     & 65.0    & 46.6     & 39.5      & 61.9      & 42.5          \\
CSWin-B \cite{CSWin}    & 97M      & 526G       & 48.7     & 70.4    & 53.9     & 43.9      & 67.8      & 47.3          \\
Pale-B \cite{Pale}      & 105M     & 595G       & 49.3     & 71.2    & 54.1     & 44.2      & 68.1      & 47.8    \\
AxWin-B         & 85M        & 370G     & \textbf{50.0}  & \textbf{71.8}  & \textbf{54.6}  & \textbf{44.7}  & \textbf{68.4}  & \textbf{48.3}   \\


\Xhline{1.0pt}
\end{tabular}
\end{table}

\begin{table}[h]
\centering
\caption{\label{downsample} Ablation study for different downsampling manner.  Flops are calculated with the resolution of 512$\times512$, note that only the down-sampling modules are tested here.}
\resizebox{0.98\linewidth}{!}{
\begin{tabular}{c|c|c|cc}
\toprule[1pt]
\multirow{2.2}{*}{Down-sampling}   & ImageNet-1K    & ADE20K    &  \\
& Top-1 acc  & SS mIoU   & \multirow{-2}{*}{Params}  & \multirow{-2}{*}{GFLOPs} \\
\midrule[1pt] 
Patch Merging\cite{ViT}    & 83.8    & 51.0    & 0.7M   & 0.6  \\
\midrule[0.5pt]
MSPE (ours)    & \textbf{83.9}   & \textbf{51.3}        & 0.7M   & 1.1   \\
\bottomrule[1pt]
\end{tabular}}
\end{table}

\noindent\textbf{Object Detection and Instance Segmentation on COCO}.
As shown in Table \ref{EXP_coco}, for object detection, our method average improves 3.2$\%$ box AP. For instance segmentation, AxWin Transformer average improves by 2.4$\%$ mask AP. Also as the input resolution increases, the average FLOPs of our method decrease by 68G.

\noindent\textbf{Semantic Segmentation on ADE20K}.
The results on ADE20K dataset are shown in Table \ref{EXP_ade20k}.
Compared to other methods, our AxWin Transformer params and FLOPs decrease more on average as the image resolution increases. Meanwhile, the performance of our single-scale mIoU and multi-scale mIoU is improved by $2.7\%$ and $2.5\%$ respectively.
\begin{table}[!h]
    \newcommand{\tabincell}[2]{\begin{tabular}{@{}#1@{}}#2\end{tabular}}
    \centering
    \caption{\label{EXP_ade20k}Comparisons of different backbones with UperNet as decoder on ADE20K for semantic segmentation. FLOPs are calculated with a resolution of $512\times2048$.}
    \resizebox{0.47\textwidth}{!}{
        \begin{tabular}{l|cc|ccc}
            \toprule[1pt]
            {Backbone}  & {Params}   &{FLOPs}      &{mIoU(SS)} &mIoU(MS)         \\
            \midrule[1pt]
            Swin-T \cite{Swin}          & 60M       & 945G    & 44.5   & 45.8 \\
            CSWin-T \cite{CSWin}        & 60M       & 959G    & 49.3   & 50.4 \\
            Pale-T \cite{Pale}          & 52M       & 996G    & 50.4   & 51.2 \\ 
            \textbf{AxWin-T} (ours)     & 52M       & 910G   & \textbf{51.3}   & \textbf{52.2}\\
            \toprule[0.5pt]
            Swin-S \cite{Swin}          & 81M       & 1038G   & 47.6   & 49.5    \\
            CSWin-S \cite{CSWin}        & 65M       & 1027G   & 50.0   & 50.8    \\
            Pale-S\cite{Pale}           & 80M       & 1135G   & 51.2   & 52.2    \\
            \textbf{AxWin-S} (ours)     & 80M       & 995G   & \textbf{52.0}   & \textbf{52.9}\\
            \toprule[0.5pt]
            Swin-B \cite{Swin}          & 121M	    & 1188G	  & 48.1  & 49.7   \\
            CSWin-B \cite{CSWin}        & 109M      & 1222G   & 50.8   & 51.7    \\
            Pale-B\cite{Pale}           & 119M      & 1311G   & 52.2  & 53.0    \\ 
            \textbf{AxWin-B} (ours)     & 97M       & 1050G   & \textbf{52.8}   & \textbf{53.7}\\
            \bottomrule[1pt]
    \end{tabular}}
\end{table}

\noindent\textbf{Ablation Study}.
Table \ref{ablation_attn_mode} compares the different attention modes and shows that our Axwin attention achieves excellent results. Table \ref{downsample} demonstrates the benefits of the MSPE module, bringing performance gains with only a small increase in computation. Table \ref{ablation_split_size} shows the performance of different split size, for tiny, small, and base models, the split size is 7, 7 and 12 respectively.

\begin{table}[h]
\centering
\caption{\label{ablation_attn_mode} Ablation study for different attention modes.}
\resizebox{0.98\linewidth}{!}{
\begin{tabular}{c|c|c|cc}
\toprule[1pt]
\multirow{2.2}{*}{Attention}   & ImageNet-1K    & ADE20K    & \multicolumn{2}{c}{COCO} \\
& Top-1 acc  & SS mIoU   & AP$^{\text{box}}$  & AP$^{\text{mask}}$ \\
\midrule[1pt] 
Axial            & 83.0    & 48.8    & 46.9   & 41.8  \\
Window     & 82.6   & 47.6    & 45.6   & 40.9  \\
\midrule[0.5pt]
AxWin (ours)    & \textbf{83.9}   & \textbf{51.3}        & \textbf{48.2}   & \textbf{43.4}   \\
\bottomrule[1pt]
\end{tabular}}
\end{table}


\begin{table}[!h]
\newcommand{\tabincell}[2]{\begin{tabular}{@{}#1@{}}#2\end{tabular}}
\centering
\caption{\label{ablation_split_size} Ablation study for different choices of split size. The padding operation is performed when the image length and width cannot be divided.}
\resizebox{0.98\linewidth}{!}{
\begin{tabular}{c|c|c|cc}
\toprule[1pt]
\multirow{2.1}{*}{\tabincell{c}{split-size\\in four stages}}   & ImageNet-1K    & ADE20K    & \multicolumn{2}{c}{COCO} \\
& Top-1 (\%)  & SS mIoU (\%)  & AP$^{\text{box}}$  & AP$^{\text{mask}}$ \\
\midrule[1pt]
\makecell[c]{3 3 3 3}   & 83.4    & 49.8    & 47.5   & 42.8  \\
\makecell[c]{5 5 5 5}   & 83.6    & 50.0    & 47.8   & 43.2  \\
\makecell[c]{\textbf{7 7 7 7}}    & \textbf{83.9}   & \textbf{51.3}   & \textbf{48.2}   & \textbf{43.4}   \\
\makecell[c]{9 9 9 9}   & 83.8    & 51.2    & 48.0   & 43.1  \\
\makecell[c]{12 12 12 12}   & 83.9    & 51.5    & 48.3   & 43.5  \\
\bottomrule[1pt]
\end{tabular}}
\end{table}

\section{Conclusion}
This work proposes an efficient self-attention mechanism, called AxWin Attention, which models both local and global context information. Based on AxWin Attention, we develop a context-aware vision transformer backbone, called AxWin Transformer, which achieves the state-of-the-art performance in ImageNet-1k image classification and outperforms previous ones in ADE20k semantic segmentation and COCO object detection and instance segmentation methods.

\clearpage
\bibliographystyle{IEEEbib}
\bibliography{strings,refs}

\end{document}